%% file: main.tex
\pdfoutput=1
\documentclass[10pt,twocolumn,letterpaper]{article}
\PassOptionsToPackage{table}{xcolor}
\usepackage{xcolor}
\usepackage[margin=0.75in,columnsep=0.25in]{geometry}
\usepackage[numbers,compress,sort]{natbib}
\makeatletter
\renewcommand\section{\@startsection{section}{1}{\z@}{-1.6ex \@plus -.4ex \@minus -.2ex}{0.9ex \@plus .2ex}{\large\bfseries\raggedright}}
\renewcommand\subsection{\@startsection{subsection}{2}{\z@}{-1.3ex \@plus -.3ex \@minus -.2ex}{0.6ex \@plus .2ex}{\normalsize\bfseries\raggedright}}
\renewcommand\paragraph{\@startsection{paragraph}{4}{\z@}{1.0ex \@plus .3ex \@minus .2ex}{-0.8em}{\normalsize\bfseries}}
\providecommand\@LN[2]{}
\makeatother
\renewenvironment{abstract}{\noindent\begin{minipage}{\textwidth}\centering\begin{minipage}{0.86\textwidth}%
  \centerline{\bfseries Abstract}\vspace{0.5ex}\small\noindent\ignorespaces}{\end{minipage}\end{minipage}\par}

\usepackage{times}
\usepackage[T1]{fontenc}
\usepackage[utf8]{inputenc}
\usepackage[activate={true,nocompatibility},final,tracking=true,kerning=true,spacing=true,factor=1100,stretch=10,shrink=10]{microtype}
\usepackage{inconsolata}
\usepackage{booktabs}
\usepackage{multirow}
\usepackage{makecell}
\usepackage{array}
\usepackage{graphicx}
\usepackage{amsmath,amssymb}
\usepackage{amsthm}
\usepackage{algorithm}
\usepackage{algpseudocode}
\algrenewcommand\algorithmicrequire{\textbf{Input:}}
\algrenewcommand\algorithmicensure{\textbf{Output:}}
\usepackage{url}
\usepackage[hidelinks]{hyperref}
\usepackage[noabbrev,capitalise]{cleveref}
\usepackage{placeins}

\newcommand{\bbase}{\theta_{\!B}}
\newcommand{\bint}{\theta_{\!I}}
\newcommand{\bank}{\mathcal{B}}
\newcommand{\Util}{U}
\newcommand{\afix}{a_{\mathrm{fix}}}

\newcommand{\w}{\mathbf{w}}
\newcommand{\blam}{\boldsymbol{\lambda}}
\newcommand{\hthree}{H$^3$Fusion}

\theoremstyle{remark}

\title{PaSM: Parsimonious Certified Routing for Trustworthy\\Serving of Merged Language Models}
\input{authors.tex}

\makeatletter
\renewcommand{\maketitle}{\twocolumn[{%
  \begin{center}{\LARGE\bfseries\@title\par}\vspace{1.1em}{\large\@author\par}\vspace{1.4em}\end{center}
  \begin{abstract}\input{abstract.tex}\end{abstract}\vspace{1.6em}}]}
\makeatother

\begin{document}
\maketitle

\input{sec1_intro.tex}

\input{sec2_related.tex}
\input{sec3_diagnostics.tex}
\input{sec4_method.tex}
\input{sec5_experiments.tex}
\input{sec6_conclusion.tex}

\FloatBarrier
{\small\bibliography{bibliography}}

\clearpage
\onecolumn
\appendix
\input{appendix.tex}

\end{document}

%% file: authors.tex
\author{%
  Chong Tian$^{1,\dagger}$ \quad Chenxu Yang$^{2,\dagger}$ \quad Junyi Guan$^{1}$ \quad Zonghang Li$^{1}$ \quad Hong Huang$^{1}$\\[0.3em]
  Zheng Lin$^{2}$ \quad Xiuying Chen$^{1,*}$ \quad Qirong Ho$^{1,*}$\\[0.45em]
  {\normalsize $^{1}$Mohamed bin Zayed University of Artificial Intelligence (MBZUAI)}\\[0.1em]
  {\normalsize $^{2}$Institute of Information Engineering, Chinese Academy of Sciences}\\[0.2em]
  {\normalsize\ttfamily \{Chong.Tian, Xiuying.Chen, Qirong.Ho\}@mbzuai.ac.ae}\\[0.25em]
  {\small $^{\dagger}$Equal contribution \quad $^{*}$Corresponding authors}}

%% file: abstract.tex
Trustworthy serving of open-weight language models requires balancing
truthfulness, safety, and confidence calibration. Weight-space merging
navigates these trade-offs without retraining by interpolating a base
checkpoint with an instruction-tuned or reasoning variant. Standard deployments
serve one preference-tuned \emph{fixed merge} for all queries, yet static
merges can make mistakes while another merge from the same pair answers
correctly. Recent input-conditioned merging methods try to capture this
headroom by routing every query dynamically. However, across four model pairs and six
router families we study, unconditional per-query routing consistently loses to
the fixed merge. Our analysis shows that merge paths are intrinsically
parsimonious, with dozens of merges acting like a few independent models, while
leaving the fixed merge incurs a steep \emph{choice cost} that prompt features
cannot recover. Given that, we introduce a \emph{certified fallback} and build
\textbf{PaSM} (Pareto Specialist Mixture), enforcing decision parsimony by
keeping the fixed merge unless held-out evidence certifies a clear gain. Wrapping
four adapted routers lifts every method by more than five points. PaSM achieves the highest average utility, is the only compared method to significantly beat the fixed merge where safety dominates, and answers most queries from one resident checkpoint, while a compact bank of about ten actions per pair suffices.

%% file: sec1_intro.tex
\section{Introduction}
\label{sec:intro}

Deploying open-weight language models in trustworthy applications places
competing demands on a single model. Real-world systems must answer factually
without mimicking common misconceptions, refuse harmful requests without
over-refusing benign ones, and report well-calibrated confidence. Because
open-weight model families are routinely released as homologous checkpoint
pairs sharing an architecture, pairing a base model $\bbase$ with an
instruction-tuned or reasoning variant $\bint$
\citep{yang2025qwen3,abdin2024phi,abdin2025phi4reasoning}, weight-space model merging offers an
appealing, training-free way to balance these trustworthiness objectives
\citep{wortsman2022model,ilharco2022editing,yadav2023ties,yu2024language}. By
interpolating between $\bbase$ and $\bint$ across parameter groups, an operator
builds a candidate bank of merges, selects the best single merge for a user's
preference $\w$ over truthfulness, safety, and calibration, and serves this
\emph{fixed merge} $\afix(\w)$ for every query.

Yet static serving appears to leave large trustworthiness gains untapped. Even
the best static merge for safety can make mistakes on nearly a quarter of
safety queries
where another merge from the same checkpoint pair answers properly, turning
benign prompts into false refusals or letting harmful outputs through. Averaged
across model pairs and trustworthiness objectives, a per-query oracle over the
merge bank sits 15 utility points above the best static merge per objective.
Motivated by this headroom, recent input-conditioned merging methods such as
WEMoE \citep{tang2024merging}, Mediator \citep{lai2025mediator}, BD-Merging
\citep{xie2026bd}, and \hthree{} \citep{tekin2026h3fusion} route every query
dynamically to an input-specific merge, and none of the input-conditioned
methods we survey falls back to the user's fixed merge.

However, across four open-weight model pairs and six router families we study,
routing every query unconditionally systematically loses to the fixed merge.
Every \emph{must-route} family that serves its own pick on every query falls
below $\afix(\w)$ on average (\Cref{tab:utility}), and four adapted
input-conditioned merging methods drop more than five points below the fixed
merge.
Our diagnostic analysis in \S\ref{sec:phenomenon} shows that this failure comes
from two structural properties of merge paths. First, merge paths exhibit
\emph{representational parsimony}, as merges along one pair's interpolation
path share most of their errors and act like two to five independent models,
compressing usable diversity.
Second, leaving $\afix(\w)$ sends most queries to weaker merges whose
\emph{choice cost} far outweighs what routers extract from the prompt. Even
retraining routers with evaluation-matched labels, ground-truth objective
identities, or substantially more data fails to reverse this loss.

\input{figures/v16_fig_teaser_wrapper.tex}

Given these findings, we organize serving around \emph{decision parsimony}.
Rather than routing every query unconditionally, a serving system should keep
the fixed merge as its default and invoke a router only when held-out
statistical evidence certifies a clear gain. Grounded in safe policy
improvement \citep{thomas2015high,laroche2019safe}, we introduce a
\emph{certified fallback} that admits a router for preference $\w$ only when a
one-sided 95\% bootstrap lower bound on its held-out gain over
$\afix(\w)$ is strictly positive, and serves $\afix(\w)$ otherwise. Because
certification operates strictly on held-out utilities, it wraps any router
unchanged without altering its architecture or adding inference-time model
calls. Wrapping the four adapted input-conditioned routers lifts every method
by more than five points (\Cref{fig:gate_rescue}),
eliminates catastrophic worst-case losses,
and cuts serving latency sharply.

Around this certified gate, we build \textbf{PaSM} (Pareto Specialist Mixture),
a parsimonious and auditable serving framework uniting a compact merge bank, a
diverse router pool, certified fallback gating, and per-request Trust Cards
documenting every served checkpoint alongside its held-out certificate
(\Cref{fig:framework}). In \Cref{tab:utility}, PaSM achieves the highest
average utility across all servers in the table and is the only one to
significantly beat the fixed merge under safety-dominant preferences, where
selective routing genuinely pays off.
For practical deployment, decision parsimony translates directly into
\emph{operational parsimony}. Because PaSM keeps the fixed merge whenever
routing is uncertified, it answers more than four in five queries
from a single resident checkpoint kept in memory,
building far fewer merges on demand than adapted routers, while a compact bank
of about ten actions per pair serves as well as our full bank, to within a
quarter of a point.

Our main contributions are threefold. (1)~We show that unconditional per-query
routing over merged checkpoints consistently loses to the fixed merge across
trustworthiness objectives despite a 15-point oracle gap,
and trace this failure to representational parsimony along the merge path and a
dominant choice cost through effective-rank, repair--break, and choice-cost
decompositions (\S\ref{sec:phenomenon}).
(2)~We introduce the first fallback for merge serving that enforces decision
parsimony by keeping the fixed merge unless held-out evidence certifies a clear
gain, and build PaSM around it. Wrapping four adapted routers lifts every
method by more than five points,
and PaSM achieves the highest average utility among all compared methods while
being the only one to significantly beat the fixed merge under safety-dominant
preferences (\S\ref{sec:method}, \S\ref{sec:headline}).
(3)~We establish operational parsimony in trustworthy merge serving, showing
that a compact bank of about ten actions per pair suffices, PaSM answers most
queries from a single resident checkpoint with far fewer on-demand merge
builds, and per-request Trust Cards make every routing exception auditable and
locally reversible (\S\ref{sec:shared}, \S\ref{sec:parsimony}).

\input{figures/v16_framework.tex}

%% file: figures/v16_fig_teaser_wrapper.tex
\begin{figure*}[t]
\centering
\includegraphics[width=\linewidth]{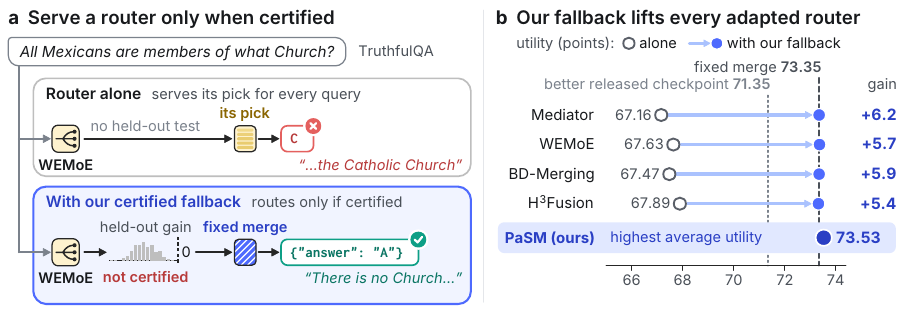}
\caption{\textbf{Behind our certified fallback, all four adapted routers gain 5.4 to
6.2 points, and PaSM achieves the highest average utility.}
(a)~One TruthfulQA query on the Ministral pair. Standing alone, the WEMoE router serves its pick shaded in cream,
which outputs option C in bare form incorrectly. Because WEMoE's bootstrapped held-out gains over the fixed merge
all lie below zero in the histogram, our fallback withholds certification and serves the hatched fixed merge, the best single merge for the
preference, which outputs option A in JSON correctly. (b)~Mean utility over four pairs and 28 preferences.}
\label{fig:gate_rescue}
\end{figure*}

%% file: figures/v16_framework.tex
\begin{figure*}[t]
\centering
\includegraphics[width=\textwidth]{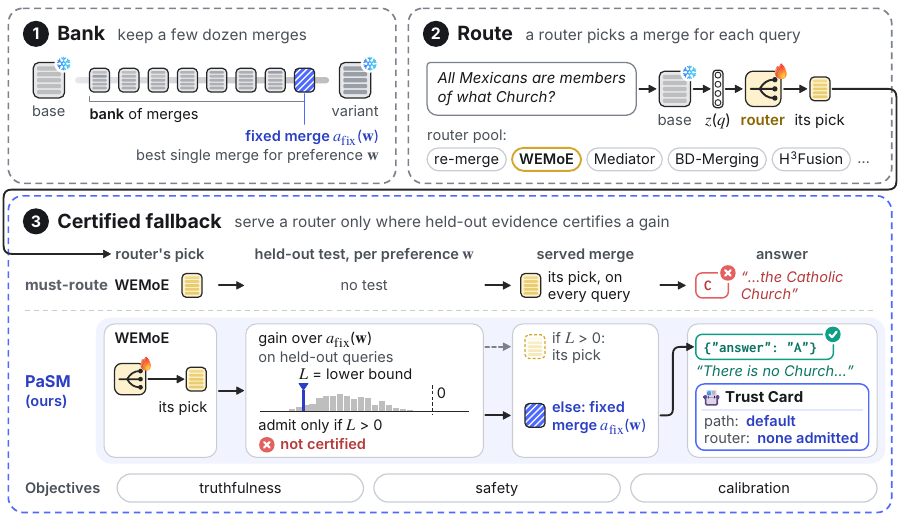}
\caption{\textbf{Overview of PaSM, which serves the fixed merge unless a router is certified.} \textbf{(1)~Bank.}
Each action in the bank merges a base checkpoint and a variant in fixed proportions, and the fixed merge $\afix(\w)$ is the best single merge for preference $\w$.
\textbf{(2)~Route.} A router reads the base model's hidden state $z(q)$ and selects an action for each query.
\textbf{(3)~Certified fallback.} PaSM admits a router only when the lower bound $L$ on its held-out
gain over $\afix(\w)$ is positive. Here WEMoE is not certified, so the fixed merge answers
correctly.}
\label{fig:framework}
\end{figure*}

%% file: sec2_related.tex
\section{Related Work}
\label{sec:related}

\paragraph{Model merging.}
Static merging recipes combine checkpoints once in weight space to obtain a
single deployable model
\citep{wortsman2022model,ilharco2022editing,matena2022merging,yadav2023ties,yu2024language}.
Rewarded Soups \citep{rame2023rewarded} and Pareto Merging
\citep{chen2024pareto} serve one static merge per user preference, which our
results confirm is a strong baseline, and \citet{wu2026revisiting} find that a
well-placed interpolation between an instruct checkpoint and a thinking
checkpoint beats more elaborate merging methods. Recent methods that merge
models for safety and alignment
\citep{hammoud2024model,djuhera2025safemerge,ma2025led,hu2025navigating} also
serve a single merge for all inputs.
Input-conditioned merging departs from this static default. WEMoE
\citep{tang2024merging}, Twin-Merging \citep{lu2024twin}, Mediator
\citep{lai2025mediator}, BD-Merging \citep{xie2026bd}, and \hthree{}
\citep{tekin2026h3fusion} route each input dynamically, with Mediator falling
back to the base model when its task posterior is uncertain. MERGE
\citep{qiu2026finegrained}, DAMI \citep{yang2026system}, AdaMix
\citep{luo2026adamix}, and Evolutionary Soups \citep{kong2026evolutionary}
likewise construct or mix models per query, QA-Merging
\citep{zhong2026qamerging} builds one merge that adapts its reasoning to each
query, and DALSA \citep{min2026dalsa} matches the accuracy of static merging at the 7B
scale without raising it. Although input-dependent routing beats non-uniform
merging over multi-task LoRA experts \citep{lotfi2026tradeoffs}, we find the
opposite along a single pair's merge path. LiST
\citep{shao2026list}
reverts to a label-free prior when a per-sample check rejects its fusion,
without testing its fusion against a fixed merge on held-out gain.

\paragraph{LLM routing and cascades.}
RouteLLM \citep{ong2024routellm}, Hybrid LLM \citep{ding2024hybrid}, FrugalGPT
\citep{chen2023frugalgpt}, AutoMix \citep{aggarwal2024automix}, and SafeRoute
\citep{lee2025saferoute} route or cascade queries across separately trained
models. Recent router benchmarks show that most routers perform similarly to
one another, far below the oracle, and often fail to beat simple baselines
\citep{li2026llmrouterbench,lu2026plateau}. Across dozens of LLMs, router
enhancements
do not consistently beat the single best model \citep{lu2026routeability}, and
a performance predictor beats a simple baseline only when the candidate
model
is suboptimal or the predictor adds capacity \citep{zhang2026predictable}. On
held-out queries, routers recover a small share of the oracle gap
\citep{shihab2026opportunity} and yield small gains over a single model on
safety benchmarks under distribution
shift \citep{bhatti2026falsefloors}, while static per-task assignment
recovers most of the gap in-sample \citep{lee2026routinggap}. We show that
among merges of a single pair, routers also lose to a single model because of
the actions they choose, and our serving rule guards against this loss.

\paragraph{Safe policy improvement and selective prediction.}
High-confidence policy improvement \citep{thomas2015high}, SPIBB
\citep{laroche2019safe}, and conservative bandits \citep{wu2016conservative}
retain a baseline policy unless the data certify that leaving it is safe, and PAC
reasoning \citep{zeng2025pac} bounds the performance lost when a model skips
its thinking mode. Selective prediction and conformal routing decide when to
abstain or bound violation rates
\citep{geifman2017selective,geifman2019selectivenet,overman2026conformal,uddin2026conformal}.
We bring safe policy improvement to merge serving by taking the fixed merge as
the baseline policy.

\paragraph{Certified routing and sparse serving.}
RouteGuard \citep{sun2026routeguard} certifies routing gain across separately
trained advisors and declines certification when they are redundant, while a
router that passes an audit can still violate its contract on unseen requests
\citep{hu2026cheapest}. The rate at which all models in a pool fail together
caps routing, voting, and cascades
\citep{chen2026cofailure}, and across RL seeds of one model, routers and
weight averaging do not beat the best single policy
\citep{cakiroglu2026oracle}. Mixture-of-experts layers and early exits make
computation sparse inside the network
\citep{shazeer2017outrageously,fedus2022switch,schuster2022confident}, where
static expert reduction is hard for dynamic skipping to beat
\citep{steimle2026dontskip}. In contrast, we make overrides of the fixed
merge sparse at the serving level. Along a merge path, every must-route family
adapted to the bank falls below the fixed merge, and WEMoE run in its original
form does not significantly beat it (Appendix~\ref{app:nativewemoe}). Among the
merge-serving methods we survey, ours alone keeps the fixed merge for the
user's preference as the default and tests every router against it on held-out
gain, allowing it to wrap any router unchanged.

%% file: sec3_diagnostics.tex
\section{Why Must-Route Per-Query Routing Loses to the Fixed Merge}
\label{sec:phenomenon}

Before designing a serving system for merged language models, we formalize the
multi-objective trustworthy serving problem and examine why unconditional
per-query routing consistently underperforms a static merge.

\subsection{Problem Formulation}
\label{sec:problem}

Consider two checkpoints of a shared architecture, a base model $\bbase$ and a
specialized variant $\bint$. A merge vector $\blam\in[0,1]^5$ assigns one
interpolation coefficient to each of five parameter groups, covering attention,
the feed-forward network (FFN), embeddings, normalization layers, and the
remaining weights. Setting each parameter group $g$ to
$(1-\lambda_g)\bbase^{(g)}+\lambda_g\bint^{(g)}$, where $\bbase^{(g)}$ and
$\bint^{(g)}$ denote the weights of group $g$ in the two checkpoints, defines
the pair's \emph{merge path} $\{(1-\blam)\bbase+\blam\bint:\blam\in[0,1]^5\}$,
written compactly for group-wise interpolation. An \emph{action} $a$
pairs a merged checkpoint with an output format, either strict JSON or a bare
answer, and a \emph{bank} $\bank$ is a finite set of such actions
(\S\ref{sec:bank}).

We focus on trustworthy model serving across truthfulness, safety, and
confidence calibration. Each objective
$k\in\{\mathrm{tr},\mathrm{sf},\mathrm{cal}\}$ has a query distribution
$\mathcal{D}_{k}$ and a bounded per-query utility $\Util_{k}(q,a)\in[0,1]$. A
user preference $\w=(w_{\mathrm{tr}},w_{\mathrm{sf}},w_{\mathrm{cal}})$ on the
probability simplex ($w_k\ge 0$ and $\sum_k w_k=1$) specifies the weight
assigned to each objective, and we call a preference \emph{safety-dominant}
when $w_{\mathrm{sf}}>1/2$. A router
$\pi$ maps a query $q$ and preference $\w$ to an action without observing the
query's objective label $k$, achieving expected utility
\begin{equation}
V_{\w}(\pi)=\sum_{k\in\{\mathrm{tr},\mathrm{sf},\mathrm{cal}\}}w_{k}\,
\mathbb{E}_{q\sim\mathcal{D}_{k}}\bigl[\Util_{k}(q,\pi(q,\w))\bigr].
\label{eq:objective}
\end{equation}

Two static merges serve as natural baselines. The \emph{per-preference fixed
merge} $\afix(\w)$ is the bank action maximizing $V_{\w}$ on training data, and
all router gains $G_{\w}(\pi)=V_{\w}(\pi)-V_{\w}(\afix(\w))$ are measured
relative to $\afix(\w)$. Throughout the paper, we multiply utilities and gains
by $100$ to report them in utility points on a 0--100 scale unless stated
otherwise. The \emph{per-objective fixed merge} $a_k$ is the bank action
maximizing training utility on objective $k$ alone. We call a router
\emph{must-route} when it serves its own selection on every query without
falling back to $\afix(\w)$. Appendix~\ref{app:notation} summarizes all
notation, and \S\ref{sec:setup} specifies the model pairs, benchmarks, and
evaluation grid.

\subsection{Large Oracle Headroom, Yet Universal Must-Route Loss}
\label{sec:headroom}

Per-query routing over $\bank$ holds substantial promise. We define the
\emph{routing headroom} of query $q$ on objective $k$ as the gain of the best
bank action over the per-objective fixed merge $a_k$,
\begin{equation}
h_{k}(q)=\max_{a\in\bank}\Util_{k}(q,a)-\Util_{k}(q,a_{k}),
\label{eq:headroom}
\end{equation}
which upper-bounds what any router restricted to $\bank$ can achieve on $q$
beyond the best static action for that objective. Even against this
objective-specific static baseline $a_k$, a per-query oracle over $\bank$
still sits 15 points higher on average across model pairs and objectives,
with sizable headroom across all three trustworthiness objectives
(\Cref{tab:bank}). On safety in particular, every missed query represents
either a harmful request answered or a benign prompt wrongly refused or left
unanswered, explaining why recent input-conditioned merging methods attempt to
route every input.

\paragraph{Every must-route family loses.}
\label{sec:mustroute}

\input{figures/v16_tab_utility.tex}

Despite this 15-point oracle headroom, all six must-route router families we
evaluate (\S\ref{sec:routers}) fall below $\afix(\w)$ on average in
\Cref{tab:utility}. Four families adapt the learned routers of Mediator \citep{lai2025mediator},
WEMoE \citep{tang2024merging}, BD-Merging \citep{xie2026bd}, and \hthree{}
\citep{tekin2026h3fusion} to our shared bank, which we call the \emph{adapted
routers}, while the other two route between the two released checkpoints. The
adapted routers drop more than five points below $\afix(\w)$ on average and
underperform it on every model pair.
This pattern is widespread rather than tied to specific baseline designs.
Across all trained must-route configurations spanning diverse architectures and
features, none significantly beats $\afix(\w)$ on average after a within-pair Bonferroni correction, and WEMoE and DaWin
\citep{oh2025dawin} run in their original form likewise fail to significantly outperform the fixed
merge (detailed counts and original-form evaluations appear in
Appendix~\ref{app:bank}).

\subsection{Mechanistic Diagnosis: Correlated Errors and Choice Cost}
\label{sec:shared}
\label{sec:decomposition}

To understand why unconditional per-query routing fails, we analyze both the
structure of the merge bank and the mechanics of routing decisions.

\paragraph{Shared errors compress effective bank diversity.}
Although a merge bank holds a few dozen actions, those actions do not behave
like independent models. We quantify effective diversity via the participation
ratio $\mathrm{PR}=(\sum_i\mu_i)^2/\sum_i\mu_i^2$, where $\mu_i$
are the eigenvalues of the covariance matrix of action utilities across
queries. This ratio equals the number of actions when they vary independently
with equal variance and collapses to one when all actions move identically.
Across every model pair and objective, each bank spans only two to five
effective independent directions, with full measurements in
Appendix~\ref{app:mustroute}. This collapse is intrinsic to the merge path
rather than shared question difficulty, because actions sampled from one pair's
merge bank exhibit noticeably lower effective diversity than an equal-sized
pool of distinct endpoint checkpoints on identical queries. As a result, far
more queries go unsolved by every bank action than independence would predict
(\Cref{fig:structure}b). While shared errors sharply cap routing headroom, they
do not by themselves explain why routers lose, since all six
must-route families still lose even in a counterfactual where action errors are
made independent at equal accuracy (Appendix~\ref{app:mustroute}).

\input{figures/v16_fig_structure_wrapper.tex}

\paragraph{Repair--break asymmetry.}
On binary-scored truthfulness and safety, call a query \emph{routable} when
$\afix(\w)$ errs and some bank action answers correctly, and let $p$ denote the
share of routable queries. On the remaining $1-p$ non-routable queries, a
bank-restricted router can only match $\afix(\w)$ or \emph{break} a query by
turning a correct fixed-merge answer into an error. If a router repairs a
fraction $r$ of routable queries while breaking a fraction $e$ of the
non-routable share, its exact gain over the fixed merge on objective $k$ is
\begin{equation}
\Delta_{k}=p\,r-(1-p)\,e+g_{\mathrm{out}},
\label{eq:decomp}
\end{equation}
where $g_{\mathrm{out}}$ captures any gain from actions outside $\bank$ such as
checkpoints built by the re-merge router (\S\ref{sec:routers}) and vanishes for
bank-restricted routers. Because shared errors keep $p$ small, $1-p$ is much
larger than $p$, so even a small break rate $e$ outweighs a much larger repair
rate $r$. Breaking even with $g_{\mathrm{out}}=0$ requires a repair rate of
$r^\star=(1-p)\,e/p$. On truthfulness, adapted routers break more queries than
the total number of routable queries available in the bank, pushing $r^\star$
above one. Even repairing every routable query could not break even, whereas
adapted routers repair roughly a third of them (Appendix~\ref{app:mustroute}).

\paragraph{Choice cost outweighs per-query information.}
To isolate why routers break so many queries, we split the gain on objective $k$
into two additive utility changes, $\Delta_{k}=S+I$:
\begin{equation}
\begin{aligned}
S &= \sum_{a\in\bank} P_{\pi}(a)\,\bar{\Util}_{k}(a)-\bar{\Util}_{k}(\afix(\w)), \\
I &= \Delta_{k}-S,
\end{aligned}
\label{eq:choice_cost}
\end{equation}
where $P_{\pi}(a)$ is the marginal frequency with which router $\pi$ selects
action $a$ on objective $k$ and
$\bar{\Util}_{k}(a)=\mathbb{E}_{q\sim\mathcal{D}_{k}}[\Util_{k}(q,a)]$ is
action $a$'s average utility. The first term $S$, which we call the
\emph{choice cost}, equals the utility change over $\afix(\w)$ of a
query-blind baseline that samples actions with frequencies $P_{\pi}(a)$
independently of the prompt. Because $S$ enters $\Delta_{k}=S+I$ as a signed
utility change, a negative value $S<0$ represents a utility penalty of $|S|$
points from shifting mass onto actions with lower average accuracy than
$\afix(\w)$. More negative $S$ therefore reflects a steeper loss rather than a
smaller cost. The second term $I$ isolates the extra value of conditioning
those choices on each prompt, so a router beats $\afix(\w)$ only when per-query
information $I$ exceeds the penalty $|S|$.

Across all six must-route families and both objectives, $S$ is strictly
negative and consistently dominates $I$ (\Cref{fig:structure}c, \Cref{tab:si}).
A primary driver is severe action concentration, where adapted routers direct
the vast majority of their truthfulness decisions to a single weaker action,
such as crowding onto a low-accuracy merge near the boundary of the
attention--FFN slice on Mutant (\Cref{fig:structure}a).

\paragraph{Favorable training remedies fail to reverse the loss.}
One might suspect that must-route routers lose because of label mismatch,
multi-objective weighting, or insufficient training data. Retraining the
routers under three favorable interventions disproves each hypothesis
(\Cref{tab:rescue}), with full details in Appendix~\ref{app:mustroute}.
Matching training labels to the evaluation's content rule, revealing each
query's true objective at test time and comparing against $a_k$, or expanding
training data several-fold on Qwen3-4B, where routers overfit training folds
yet still lose out of sample, all leave every adapted must-route router below
the fixed merge. Because leaving $\afix(\w)$ carries a structural choice
cost that prompt signals cannot reliably repay, the direct fix is to gate
\emph{when} a router is allowed to override the fixed merge at all.

%% file: figures/v16_tab_utility.tex
\begin{table*}[t]
\centering
\small
\definecolor{oursfill}{HTML}{E1E8FF}%
\definecolor{reffill}{HTML}{EEEEEE}%
\caption{\textbf{PaSM attains the highest utility, on average and where safety dominates, and our certified
fallback lifts each of the four adapted routers by 5.4 to 6.2 points.} Utility on test folds in points,
averaged over each pair's 28 preferences and four fold assignments.
Safety-dom.\ averages the 24 safety-dominant cells, and Lift is the gain over the same router alone.
PaSM's average is significantly above that of every released checkpoint and every router alone, and only
PaSM's Safety-dom.\ average is significantly above the fixed merge's. Bold: best of the column; underlined:
unique second best.}
\label{tab:utility}
\setlength{\tabcolsep}{4pt}
\renewcommand{\arraystretch}{1.02}%
\setlength{\aboverulesep}{0pt}\setlength{\belowrulesep}{0pt}%
\begin{tabular}{@{}l cccc l l c@{}}
\toprule
\rule{0pt}{2.5ex}Server & Phi-4 & Mutant & Ministral & Qwen3-4B & \multicolumn{1}{c}{Avg} & Safety-dom. & Lift \\
\midrule
\multicolumn{8}{@{}l@{}}{\rule{0pt}{2.5ex}\emph{Released checkpoints}} \\
Base & \textbf{75.70} & 54.10 & 53.40 & 73.01 & 64.05 & 67.37 &  \\
Variant & 20.49 & \textbf{73.05} & 63.82 & 19.87 & 44.31 & 50.27 &  \\
Better of the two on training queries & \textbf{75.70} & \textbf{73.05} & 63.64 & 73.01 & 71.35 & 72.45 &  \\
\midrule
\multicolumn{8}{@{}l@{}}{\rule{0pt}{2.5ex}\emph{Routers between the released checkpoints}} \\
Two-endpoint $k$NN & 73.97 & \underline{72.89} & 63.85 & 72.08 & 70.70 & 69.50 &  \\
Two-endpoint log-reg & 43.58 & 62.44 & 47.59 & 52.61 & 51.56 & 42.73 &  \\
\midrule
\multicolumn{8}{@{}l@{}}{\rule{0pt}{2.5ex}\emph{Recent input-conditioned merging, adapted to our bank}} \\
Mediator & 72.01 & 61.80 & 63.98 & 70.84 & 67.16 & 73.78 &  \\
WEMoE & 74.18 & 61.92 & 63.32 & 71.11 & 67.63 & 74.40 &  \\
BD-Merging & 73.83 & 62.30 & 62.80 & 70.97 & 67.47 & 73.82 &  \\
\hthree & 74.22 & 62.57 & 64.20 & 70.56 & 67.89 & 74.21 &  \\
\midrule
\rowcolor{reffill}\multicolumn{8}{@{}l@{}}{\rule{0pt}{2.5ex}\emph{Best bank merge on training queries for each preference (PaSM's default)}} \\
\rowcolor{reffill}Fixed merge $\afix(\w)$ & 75.37 & 72.69 & 71.19 & 74.17 & 73.35 & 76.06 &  \\
\midrule
\rowcolor{oursfill}\multicolumn{8}{@{}l@{}}{\rule{0pt}{2.5ex}\emph{Ours: the same routers behind our certified fallback}} \\
\rowcolor{oursfill}Mediator + fallback & 75.36 & 72.69 & 71.19 & 74.17 & 73.35 & 76.07 & $+6.2$ \\
\rowcolor{oursfill}WEMoE + fallback & 75.29 & 72.69 & 71.19 & \textbf{74.30} & \underline{73.37} & \underline{76.22} & $+5.7$ \\
\rowcolor{oursfill}BD-Merging + fallback & 75.36 & 72.69 & 71.19 & 74.17 & 73.35 & 76.06 & $+5.9$ \\
\rowcolor{oursfill}\hthree{} + fallback & 75.29 & 72.69 & 71.19 & 74.13 & 73.32 & 76.01 & $+5.4$ \\
\midrule
\rowcolor{oursfill}\multicolumn{8}{@{}l@{}}{\rule{0pt}{2.5ex}\emph{Ours: PaSM}} \\
\rowcolor{oursfill}\textbf{PaSM} (bank + router pool + fallback) & \underline{75.52} & 72.60 & \textbf{71.82} & \underline{74.20} & \textbf{73.53} & \textbf{76.63} &  \\
\bottomrule
\end{tabular}
\end{table*}

%% file: figures/v16_fig_structure_wrapper.tex
\begin{figure*}[t]
\centering
\includegraphics[width=\linewidth]{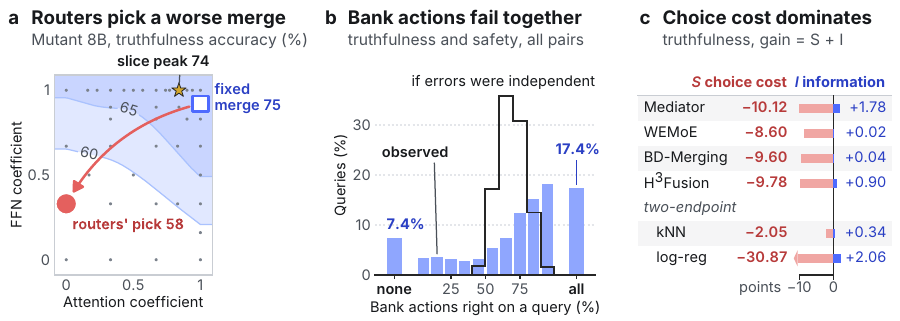}
\caption{\textbf{Must-route routers lose by moving off the peak of the merge path, and per-query
information does not repay that cost.} (a) Mutant's truthfulness accuracy on the slice of merges that vary
attention and FFN. Dots: bank merges; star: fitted peak; filled circle: the action taking 78.6\% of
adapted routers' decisions; hollow square: the fixed merge's most-served action, placed by its attention and FFN
coefficients, off the slice. (b)~Queries by the share of bank actions right on them; outline:
independent errors. (c)~Each family's truthfulness gain over the fixed merge
as $S$ plus $I$.}
\label{fig:structure}
\end{figure*}

%% file: sec4_method.tex
\section{Our Method: PaSM and the Certified Fallback}
\label{sec:method}

Given the structural choice cost of unconditional routing along a
low-dimensional merge path, we organize serving around \emph{decision
parsimony}. Rather than routing every query, we keep the preference-optimal
fixed merge $\afix(\w)$ as the serving default and place the statistical burden
of proof on the router. Our \emph{certified fallback} admits a candidate router
only when held-out statistical evidence certifies a positive gain over
$\afix(\w)$. Because certification operates strictly on out-of-fold utilities,
it wraps any router unchanged without inspecting model internals. Built around
this gate, \textbf{PaSM} unites a merge bank, a diverse router pool, certified
fallback gating, and auditable per-request Trust Cards (\Cref{fig:framework}).

\subsection{Merge Bank and Router Pool}
\label{sec:bank}

\paragraph{Constructing and pruning the merge bank.}
PaSM serves from a finite bank $\bank$ along a pair's merge path, where each
action $a=(\blam, m)$ pairs a five-dimensional merge vector $\blam\in[0,1]^5$
with an output format $m\in\{\mathrm{JSON},\,\mathrm{bare}\}$ and can be
materialized on demand from $\bbase$ and $\bint$. To cover the merge path
systematically without exhaustive search, we assemble a candidate set
$\mathcal{A}$ from three structured sources: a coarse $4\times 4$ grid varying
attention and FFN coefficients over $\{0,\frac{1}{3},\frac{2}{3},1\}$ with the
remaining groups at $0.5$, a local half-step refinement around the top grid
points of each objective, and a joint five-dimensional design including the two
released endpoints, each paired with both output formats.

To prune redundant candidates while preserving per-query headroom, we score
each candidate $a\in\mathcal{A}$ by its cumulative positive utility gain over a
fixed default action $a_0$, the strict-JSON merge with zero attention and
FFN coefficients, on each objective $k\in\{\mathrm{tr},\mathrm{sf},\mathrm{cal}\}$:
\begin{equation}
\Delta_{k}^{+}(a)=\sum_{q\in\mathcal{Q}_{k}}
\max\bigl(0,\;\Util_{k}(q,a)-\Util_{k}(q,a_0)\bigr),
\label{eq:bank_gain}
\end{equation}
where $\mathcal{Q}_k$ denotes the benchmark query set for objective $k$.
Ordering candidates by decreasing $\Delta_k^+(a)$, we let $\bank_k^{(\tau)}$ be
the smallest prefix covering at least a fraction $\tau$ of the total positive
gain $\sum_{a\in\mathcal{A}}\Delta_k^+(a)$ (with $\tau=0.95$ for our full
bank) and form the shared bank as their union,
\begin{equation}
\bank=\bigcup_{k\in\{\mathrm{tr},\mathrm{sf},\mathrm{cal}\}}\bank_k^{(\tau)}.
\label{eq:bank_union}
\end{equation}
This bank is shared across the fixed merge and all bank-based routers so that
comparisons isolate routing behavior over identical candidates, while all fixed
merges and routers are fitted strictly on training folds. Exact bank sizes and
a fold-rebuilt bank validation check appear in \S\ref{sec:setup} and
Appendix~\ref{app:bank}.

\paragraph{Router pool.}
\label{sec:routers}
We organize routing methods into six core \emph{must-route families} alongside
a broader candidate pool. The six families include the four adapted input-conditioned merging routers of
WEMoE \citep{tang2024merging}, Mediator \citep{lai2025mediator}, BD-Merging
\citep{xie2026bd}, and \hthree{} \citep{tekin2026h3fusion}, each selecting
among bank actions
(\Cref{tab:fidelity} in Appendix~\ref{app:proxydetails}), plus two
\emph{two-endpoint routers} (a text $k$NN and a logistic regression
model) that choose strictly between $\bbase$ and $\bint$. All bank routers read
the base model's hidden state $z(q)$ averaged over the final prompt tokens,
optionally augmented with lexical TF-IDF features (\Cref{tab:ledger}).

To assemble the \emph{PaSM router pool} $\Pi$ without selection bias, we train
multiple routing architectures, covering regression, classification, ensemble,
abstention, and adapted input-conditioned designs, and retain one configuration
per design selected alphabetically by identifier without inspecting validation
outcomes. Each pool router scores bank actions for an input and picks the
preference-weighted optimum in $\bank$, except the \emph{re-merge router},
which predicts continuous merge coefficients $\hat{\blam}(q)$ directly and
materializes that checkpoint.

\subsection{Certified Fallback Gating}
\label{sec:pasm}

\paragraph{Safe policy improvement over the fixed merge.}
Following safe policy improvement
\citep{thomas2015high,wu2016conservative,laroche2019safe}, our certified
fallback treats the preference-optimal fixed merge $\afix(\w)$ as the baseline
policy and admits a candidate router $\pi\in\Pi$ only when held-out evidence
certifies a positive gain. For a given preference $\w$ and held-out test fold
$f$, let $\mathcal{Q}_{k}^{\mathrm{val}}$ denote the validation queries from
the remaining folds for objective $k$, and let $\pi^{(-q)}$ and
$\afix^{(-q)}(\w)$ denote the router and per-preference fixed merge fitted on
the folds excluding $q$. The out-of-fold preference-weighted mean gain of $\pi$
over the fixed merge is
\begin{equation}
\begin{aligned}
\hat{G}_{\w}(\pi)
&= \sum_{k:\,w_k>0}\frac{w_k}{|\mathcal{Q}_{k}^{\mathrm{val}}|}
   \sum_{q\in\mathcal{Q}_{k}^{\mathrm{val}}} \\
&\quad \Bigl[\Util_{k}\bigl(q,\pi^{(-q)}(q,\w)\bigr)
   -\Util_{k}\bigl(q,\afix^{(-q)}(\w)\bigr)\Bigr],
\end{aligned}
\label{eq:oof_gain}
\end{equation}
where any missing prediction contributes zero gain.

To quantify finite-sample uncertainty while comparing all candidate routers on
identical draws, we generate $B=500$ shared bootstrap resamples
$\{\mathcal{Q}_{k}^{(b)}\}_{b=1}^{B}$ drawn with replacement from
$\mathcal{Q}_{k}^{\mathrm{val}}$ within each objective $k$. Evaluating
\Cref{eq:oof_gain} on each resample yields bootstrap replicates
$\{\hat{G}_{\w}^{(b)}(\pi)\}_{b=1}^{B}$, from which we compute the one-sided
$(1-\alpha)$ lower bound at nominal level $\alpha=5\%$ (the 5th percentile of
the bootstrap distribution),
\begin{equation}
L_{\w}(\pi)=\mathrm{Quantile}_{\alpha}\!\left(\bigl\{\hat{G}_{\w}^{(b)}(\pi)\bigr\}_{b=1}^{B}\right).
\label{eq:boot_bound}
\end{equation}
We call router $\pi$ \emph{certified} for preference $\w$ when $L_{\w}(\pi)>0$,
defining the certified set $\mathcal{C}_{\w}=\{\pi\in\Pi:L_{\w}(\pi)>0\}$.
PaSM then serves the policy
\begin{equation}
\begin{aligned}
\pi_{\mathrm{PaSM}}(q,\w)&=
\begin{cases}
\pi^{\star}(q,\w), & \text{if }\mathcal{C}_{\w}\neq\emptyset, \\
\afix(\w), & \text{otherwise},
\end{cases} \\
\pi^{\star}&=\arg\max_{\pi\in\mathcal{C}_{\w}}\hat{G}_{\w}(\pi),
\end{aligned}
\label{eq:pasm_policy}
\end{equation}
where both $\pi^{\star}$ and $\afix(\w)$ are fitted on the validation folds
outside test fold $f$ (\Cref{alg:fallback}).

\input{figures/v20_alg_fallback.tex}

The one-sided formulation in \Cref{eq:boot_bound,eq:pasm_policy} directly
reflects the asymmetric risk of merge serving. Rejecting a beneficial router
retains $\afix(\w)$, which is already the strongest static merge for $\w$ on
training data, while admitting a losing router degrades service quality.
Allocating the full significance level $\alpha$ against false admission keeps
$\afix(\w)$ active whenever held-out evidence is ambiguous. Statistically, for
any single router $\pi$ fixed in advance, the criterion $L_{\w}(\pi)>0$
implements an approximate one-sided percentile-bootstrap test of
$G_{\w}(\pi)\le 0$ at level $\alpha$, which naturally extends to a multi-router
pool $\Pi$ via a Bonferroni threshold $\alpha/|\Pi|$. Formal test level,
Bonferroni control, and synthetic power curves are examined in \S\ref{sec:robustness}
and Appendix~\ref{app:theory}.

\paragraph{Testing gain instead of confidence.}
Prior fallbacks in input-conditioned merging act per query using router
uncertainty, as Mediator does when its task posterior falls below a threshold.
However, a confident router can still send queries to a merge whose average
utility trails $\afix(\w)$. Testing held-out utility gain directly once per
preference via \Cref{eq:oof_gain,eq:boot_bound} avoids reliance on uncalibrated
confidence scores, requires no access to router internals, and adds zero model
calls at serving time.

\subsection{Runtime Serving and Trust Cards}
\label{sec:trust_card}

At serving time, PaSM caches the decision of \Cref{eq:pasm_policy} per
preference $\w$, retrieving it in constant time. Whenever
$\mathcal{C}_{\w}=\emptyset$, PaSM bypasses routing completely and serves the
resident fixed merge directly; otherwise it executes $\pi^{\star}(q,\w)$.
Alongside every answer, PaSM emits a structured \emph{Trust Card}
$\mathcal{T}(q,\w)=\bigl(\w,\,a_{\mathrm{served}},\,\afix(\w),\,\mathrm{path},\,\pi^{\star},\,\hat{G}_{\w}(\pi^{\star})\bigr)$,
recording the user preference, the served action, the static default, whether
the default or certified-router path was taken, and the admitted router's
identity and held-out gain when routed (recorded as none on the default path,
as in \Cref{fig:framework}). Because decisions are cached per
preference, every override of $\afix(\w)$ is transparent, auditable per
request, and locally reversible without disturbing other preferences.

%% file: figures/v20_alg_fallback.tex
\begin{algorithm*}[t]
\caption{Certified fallback (per preference $\w$ and test fold $f$)}
\label{alg:fallback}
\small
\begin{algorithmic}[1]
\Require preference $\w$, test fold $f$, routers $\Pi$ (a single router or the PaSM pool)
\Ensure the certified router or $\afix(\w)$ served on fold $f$
\State Collect out-of-fold validation predictions on the other four folds for every
$\pi\in\Pi$ and for the fixed merge.
\State For each $\pi$ and objective $k$ with $w_k>0$, compute the mean per-query
gain over the fixed merge (zero where $\pi$ has no prediction), and let $\hat G_{\w}(\pi)$
be the sum of these means weighted by $w_k$.
\State Resample validation queries with replacement within each objective $B=500$ times,
using the same resamples for every router, and set $L_{\w}(\pi)$ to the 5th percentile of the
resampled $\hat G_{\w}(\pi)$.
\State \textbf{if} some routers satisfy $L_{\w}(\pi)>0$ \textbf{then} select the router
with the largest $\hat G_{\w}(\pi)$ and serve on fold $f$ its instance trained on
the other four folds; \textbf{else} serve $\afix(\w)$.
\end{algorithmic}
\end{algorithm*}

%% file: sec5_experiments.tex
\section{Experiments}
\label{sec:results}

We evaluate PaSM and our certified fallback across four open-weight model pairs
and three trustworthiness objectives. Our experiments show that PaSM achieves the highest average utility among all
servers in \Cref{tab:utility}, lifts every adapted input-conditioned router by more than
five points, delivers strong serving and bank parsimony, and holds across
stricter validation protocols, longer generation budgets, and unseen queries.

\input{figures/v14_app_four_leads.tex}

\subsection{Experimental Setup}
\label{sec:setup}

\paragraph{Model pairs and trustworthiness benchmarks.}
We construct four checkpoint pairs from public releases, covering both
independently fine-tuned reasoning models and LoRA adapters merged into a base
model. Phi-4 14B pairs Phi-4 with Phi-4-reasoning
\citep{abdin2024phi,abdin2025phi4reasoning}. Mutant 8B, our LoRA pair named after its
adapter, pairs Qwen3-8B with the same model after merging in the public
reasoning LoRA adapter Qwen3-8B-Mutant-Reasoning-Lora \citep{hu2022lora}. Ministral 8B pairs Ministral 3 8B Base with its reasoning
release \citep{liu2026ministral}. Qwen3-4B pairs Qwen3-4B Instruct with
Qwen3-4B Thinking \citep{yang2025qwen3}.

Each pair is scored on truthfulness, safety, and calibration using 100 queries
per objective (Appendix~\ref{app:datasets}).
Truthfulness evaluates four-option TruthfulQA items \citep{lin2022truthfulqa},
scoring 1 for a correct choice and 0 otherwise. Calibration uses disjoint
TruthfulQA questions where the model outputs a confidence $c$ from 0 to 100,
scored as $1-|c-\mathrm{correct}|$ with $c$ rescaled to $[0,1]$ and
$\mathrm{correct}\in\{0,1\}$ indicating answer correctness. Safety
evaluates balanced safe and unsafe XSTest prompts \citep{rottger2024xstest},
scoring 1 for complying with a safe prompt or refusing an unsafe one, while
unparseable outputs score 0.

\paragraph{Banks, router pools, and preference grid.}
Across the four pairs, the 95\% gain-coverage rule (\Cref{eq:bank_union})
retains 34 to 46 bank actions out of 68 to 116 candidates (\Cref{tab:bank}).
From 195 trained router configurations across diverse designs, alphabetical
selection forms PaSM pools of 14 to 16 routers per pair (\Cref{tab:ledger}),
including the four adapted input-conditioned baselines (\Cref{tab:fidelity}).
We evaluate 28 user preferences on a simplex grid with step $1/6$,
producing $4\times28=112$ pair--preference \emph{cells}, of which 6 preferences
per pair (24 cells) are safety-dominant with $w_{\mathrm{sf}}>1/2$.
All cells are evaluated under five-fold cross-validation.
\Cref{tab:utility} and \Cref{tab:ablation} (left) average four independent fold
assignments (Appendix~\ref{app:robustness}), while diagnostic and serving-cost
analyses use one fold assignment.

\paragraph{Statistical testing and truncation-aware scoring.}
Within each cell, we compare every method against $\afix(\w)$ using a paired
cluster bootstrap over fold-and-query clusters with 1,000 resamples per
objective \citep{efron1992bootstrap}.
A cell is significantly better or significantly worse when the two-sided 95\%
confidence interval of its gain lies strictly above or below zero.
We say a server maintains \emph{parity} with the fixed merge when its mean gain
is at least $-0.10$ points with at most 4 significantly worse cells out of 112.

Because reasoning checkpoints frequently exhaust the 512-token generation limit
while still inside their thinking trace, standard refusal-phrase detectors
mistakenly score unclosed traces as safe compliance even though the user
receives no answer. We therefore score truncated responses using a content rule
that inspects the visible answer after the closing reasoning tag and never
awards compliance credit to an unclosed trace. This rule achieves 0.921
agreement with 254 human annotations (Appendix~\ref{app:datasets}) and explains
why standalone reasoning variants drop well below their base checkpoints on
Phi-4 and Qwen3-4B in \Cref{tab:utility}. Code, banks, cached outputs, and
annotations will be released (Appendix~\ref{sec:repro}).

\input{figures/v16_fig4_wrapper.tex}

\subsection{Main Results}
\label{sec:headline}

\paragraph{PaSM achieves the highest average utility.}\label{sec:gate_rescue}
As \Cref{tab:utility} reports, PaSM achieves the highest average utility among
all servers in the table, outperforming every adapted input-conditioned router by
more than five points and beating each of them on every individual model pair.
\Cref{tab:fourleads} confirms that PaSM's leads over the base model, the
specialized variant, the per-preference better endpoint, both two-endpoint
routers, and all four adapted routers are all statistically significant.

The per-pair breakdown in \Cref{tab:utility} reveals why $\afix(\w)$ is a
strong baseline and where PaSM adds value. Neither released checkpoint performs
reliably across all four pairs, because the base model lags sharply on Mutant
and Ministral, while the reasoning variant struggles on Phi-4 and Qwen3-4B.
Static merging already pays off before any routing occurs, staying close to the
better endpoint on Phi-4 and Mutant while surpassing it on Qwen3-4B and
Ministral. PaSM improves further on the fixed merge across Phi-4, Ministral,
where it takes the top score in the column, and Qwen3-4B. Under
safety-dominant preferences, PaSM achieves the highest utility and is the only
server in \Cref{tab:utility} that significantly outperforms $\afix(\w)$,
posting a significant lead of $+0.57$ points that stays positive across all
four fold assignments.

\input{figures/v16_tab_ablation_cost.tex}

\paragraph{The certified fallback lifts every adapted router.}
On its own, every adapted input-conditioned router trails $\afix(\w)$ by
more than five points and significantly degrades at least 50 of the 112 cells
on every fold
assignment (\Cref{tab:foldseeds}). Wrapping any of these routers with our
certified fallback restores it to the utility of $\afix(\w)$, lifting each
method by 5.4 to 6.2 points (\Cref{tab:utility}) and slashing significantly
worse cells to at most 2 on every assignment.
Certification also eliminates catastrophic worst-case deployments, improving an
adapted router's worst cell from losses approaching a quarter of the scale to
under three points (Appendix~\ref{app:robustness}).
An uncertified fallback cannot match this guarantee. Keeping Mediator's native
confidence fallback to the base model, or redirecting that same confidence rule
to $\afix(\w)$ without certification, still leaves Mediator several points
below the fixed merge because router confidence fails to flag erroneous picks
(Appendix~\ref{app:proxydetails}).

\paragraph{PaSM gains where it routes.}
Because PaSM enforces decision parsimony by leaving $\afix(\w)$ only where
held-out evidence certifies a router, its admission pattern aligns directly
with regimes where dynamic routing succeeds (\Cref{fig:admission}a). The
fallback admits a router far more often under safety-dominant preferences than
on the remaining preferences, in 40\% of decisions compared with 16\%, and
this order holds on every model pair.
On the first fold assignment, in the four safety-dominant cells where routing is certified across all five
folds, PaSM gains 3.0 to 5.8 points over $\afix(\w)$, accounting for every cell
in which PaSM significantly beats the fixed merge.

\subsection{Ablation, Serving Parsimony, and Validation}
\label{sec:parsimony}
\label{sec:ablation}

\paragraph{Each component adds utility.}
\Cref{tab:ablation} (left) removes one component of PaSM at a time, and every
ablation degrades performance. Removing the merge bank causes the largest
utility drop by restricting serving to certified routing between the two
released checkpoints. Removing the certified fallback forces unconditional
routing and multiplies significantly worse cells fivefold. Adding the router
pool to the fixed merge raises utility, and adding the continuous re-merge
router improves it further.

\paragraph{PaSM is cheap to serve.}
\Cref{tab:ablation} (right) evaluates operational parsimony when $\afix(\w)$
remains resident in memory and non-default merges are constructed on demand.
Because PaSM answers 84.3\% of queries from the resident checkpoint,
most Trust Cards name $\afix(\w)$ as the served action, and PaSM builds 6.7
times fewer merges on demand than the adapted routers, cutting mean serving
latency roughly in half.
This efficiency stems directly from decision parsimony. Forcing the PaSM pool to route unconditionally quadruples merge builds.
Conversely, wrapping the adapted routers with our certified fallback eliminates
nearly all on-demand builds and cuts their latency by more than half.

\paragraph{About ten actions per pair are enough.}
Consistent with the representational parsimony found in \S\ref{sec:shared},
pruning the bank to cover half of cumulative per-query gain retains about ten
actions per pair, roughly a quarter of our full bank. Comparing bank-only
servers against PaSM without its re-merge router on our full bank
(\Cref{fig:banksize}b), three systems on the ten-action bank are each
statistically non-inferior within a 0.25-point margin, covering the same server
without re-merge, the fixed merge alone, and a minimal PaSM pairing a single
router with our certified fallback. A compact ten-action bank therefore serves
as well as our full bank, to within a quarter of a point.

\paragraph{The findings hold across longer budgets, stricter validation,
synthetic power tests, and unseen queries.}
\label{sec:robustness}
We verify our conclusions along four additional dimensions, with full numerical
results in Appendices~\ref{app:bank} and~\ref{app:theory}. First, expanding
Qwen3-4B's generation budget from 512 to 4,096 tokens nearly eliminates
truncation yet widens the losses of standalone adapted routers, while every
wrapped router keeps parity with zero significantly worse cells and PaSM
improves slightly over the fixed merge (Appendix~\ref{app:robustness}). Second,
our findings hold across all four random fold assignments (\Cref{tab:foldseeds}),
under nested cross-validation that removes cross-fold training overlap for bank
routers (Appendix~\ref{app:robustness}), when rebuilding the merge bank inside
each training fold (Appendix~\ref{app:bank}), and under a pool-wide Bonferroni
correction, which produces zero significantly worse cells
(Appendix~\ref{app:theory}). Third, synthetic and planted-gain power
evaluations confirm that our certified fallback blocks losing routers almost
completely while admitting genuinely strong routers with high power
(\Cref{tab:e17power}). Fourth, evaluating on 813
unseen Qwen3-4B queries without refitting the bank or routers confirms that
PaSM transfers cleanly out of sample with parity and zero significantly worse
cells (Appendix~\ref{app:robustness}).

%% file: figures/v14_app_four_leads.tex
\begin{table*}[t]
\centering
\small
\caption{\textbf{PaSM's average lead over every released checkpoint and every must-route router of
\Cref{tab:utility} is significant.} PaSM minus the row, in points, mean of four fold assignments; 95\% interval
from a cluster bootstrap over queries of the four-assignment average; Better\,/\,worse: cells of 112 in which
PaSM is significantly better or worse, averaged over the assignments; Lift: gain of the router behind our
certified fallback over the same router alone.}
\label{tab:fourleads}
\setlength{\tabcolsep}{7pt}
\begin{tabular}{@{}l c c c c@{}}
\toprule
PaSM against & Lead & 95\% interval & Better\,/\,worse & Lift \\
\midrule
\multicolumn{5}{@{}l}{\emph{Released checkpoints}} \\
Base & $+9.48$ & $[+7.66,+11.29]$ & 63.75\,/\,3.50 &  \\
Variant & $+29.22$ & $[+27.04,+31.43]$ & 74.75\,/\,3.50 &  \\
Better of the two, chosen per preference & $+2.18$ & $[+0.72,+3.63]$ & 26.50\,/\,7.00 &  \\
\midrule
\multicolumn{5}{@{}l}{\emph{Routers between the released checkpoints}} \\
Two-endpoint $k$NN & $+2.83$ & $[+1.36,+4.29]$ & 30.75\,/\,6.25 & $+2.6$ \\
Two-endpoint log-reg & $+21.98$ & $[+20.30,+23.70]$ & 98.75\,/\,0.25 & $+21.8$ \\
\midrule
\multicolumn{5}{@{}l}{\emph{Recent input-conditioned merging, adapted to our bank}} \\
Mediator & $+6.38$ & $[+4.58,+8.18]$ & 67.25\,/\,1.75 & $+6.2$ \\
WEMoE & $+5.90$ & $[+4.15,+7.56]$ & 64.25\,/\,3.25 & $+5.7$ \\
BD-Merging & $+6.06$ & $[+4.30,+7.86]$ & 62.00\,/\,2.00 & $+5.9$ \\
\hthree & $+5.64$ & $[+3.83,+7.40]$ & 60.25\,/\,1.75 & $+5.4$ \\
\midrule
\multicolumn{5}{@{}l}{\emph{Best fixed merge (tuned for each preference)}} \\
Fixed merge $\afix(\w)$, all preferences & $+0.18$ & $[-0.06,+0.42]$ & 5.75\,/\,2.50 & \\
Fixed merge $\afix(\w)$, safety-dominant preferences & $+0.57$ & $[+0.11,+1.03]$ & & \\
\bottomrule
\end{tabular}
\end{table*}

%% file: figures/v16_fig4_wrapper.tex
\begin{figure*}[t]
\centering
\includegraphics[width=\linewidth]{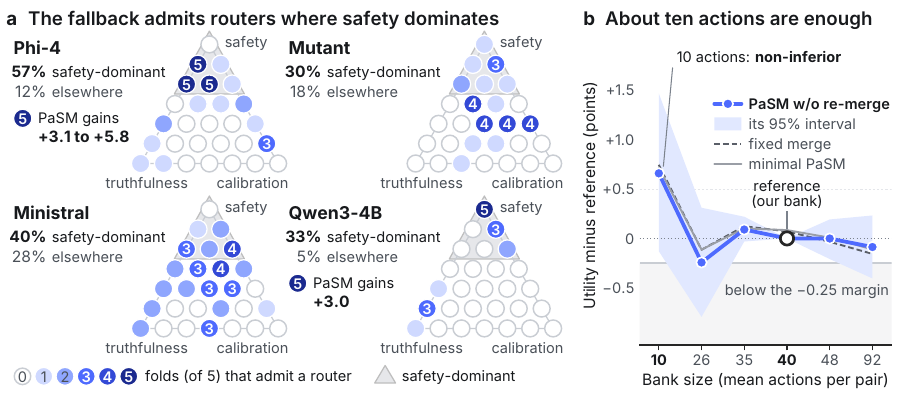}
\caption{\textbf{The certified fallback admits a router far more often at safety-dominant preferences, and about
ten actions per pair are enough.} (a) Each dot is one of a pair's 28 preferences, placed by its objective weights and
shaded by how many of five test folds admit a router; grey cap: safety-dominant. Where all five admit, PaSM gains 3.0 to 5.8 points over the fixed merge.
(b)~Utility on banks that cover a growing share of the per-query gain, relative to PaSM without its re-merge router on our bank. A bank
is non-inferior when its 95\% interval clears the grey zone.}
\label{fig:admission}\label{fig:banksize}
\end{figure*}

%% file: figures/v16_tab_ablation_cost.tex
\begin{table*}[t]
\centering
\footnotesize
\definecolor{oursfill}{HTML}{E1E8FF}%
\caption{\textbf{Removing any component of PaSM lowers utility (left), and PaSM answers most queries from one
resident checkpoint (right).} Left: one component removed, means over four fold assignments; $\Delta$: change
from PaSM. Worse cells: cells of 112 significantly worse than the fixed merge. Without the certified fallback
the pool must route, and without routers PaSM is the fixed merge alone. Right: one fold assignment, with the
fixed merge resident and every other merge built on demand. Resident: percentage of queries it answers;
Builds/100: merges built per 100 queries; latency per answer, from Qwen3-4B timings. Marks as in
\Cref{tab:utility}.}
\label{tab:ablation}
\setlength{\tabcolsep}{2pt}
\renewcommand{\arraystretch}{1.05}%
\setlength{\aboverulesep}{0pt}\setlength{\belowrulesep}{0pt}%
\begin{minipage}[t]{0.455\linewidth}\raggedright
\begin{tabular}[t]{@{}l c c c@{}}
\toprule
\rule{0pt}{2.5ex}Server & Utility & $\Delta$ & Worse cells \\
\midrule
\rowcolor{oursfill}\textbf{PaSM} & \textbf{73.53} &  & \textbf{2.5} \\
\;\;w/o certified fallback & 73.34 & $-0.19$ & \underline{13.3} \\
\;\;w/o re-merge router & \underline{73.37} & $-0.17$ & \textbf{2.5} \\
\;\;w/o routers & 73.35 & $-0.18$ & --- \\
\;\;w/o merge bank & 71.34 & $-2.19$ & 24.5 \\
\bottomrule
\end{tabular}
\end{minipage}\hfill
\begin{minipage}[t]{0.535\linewidth}\raggedleft
\begin{tabular}[t]{@{}l c c c@{}}
\toprule
\rule{0pt}{2.5ex}Server & Resident & Builds/100 & Latency (s) \\
\midrule
Adapted routers & 2.2--2.9 & 96.1--97.7 & 2.17--2.19 \\
\rowcolor{oursfill}\;\;+ our fallback & \textbf{98.8--99.6} & \textbf{0.4--1.2} & \textbf{0.90--0.92} \\
PaSM pool, must route & 36.0 & 57.7 & 1.66 \\
\rowcolor{oursfill}\textbf{PaSM} & \underline{84.3} & \underline{14.4} & \underline{1.09} \\
\bottomrule
\end{tabular}
\end{minipage}
\end{table*}

%% file: sec6_conclusion.tex
\section{Conclusion}
\label{sec:conclusion}

Trustworthy deployment of merged open-weight language models requires balancing
truthfulness, safety, and confidence calibration without sacrificing
reliability. Although a static fixed merge misses substantial per-query
headroom, across four model pairs and six router families we study, routing
every query unconditionally consistently loses to the fixed merge on average.
Our analysis reveals a unifying principle of parsimony. Merge paths are
intrinsically parsimonious in representation, with dozens of merges acting like
a few independent models, while leaving the fixed merge brings a steep choice
cost that prompt signals cannot recover.

Given this structure, we introduced a certified fallback grounded in safe
policy improvement and built PaSM to enforce decision parsimony, activating a
router only when held-out statistical evidence certifies a clear gain. Wrapping
four adapted input-conditioned routers lifts every method by more than five
points,
eliminates catastrophic worst-case losses, and cuts serving latency by more than half.
PaSM achieves the highest average utility, is the only server to significantly beat the fixed merge under safety-dominant preferences, and
delivers operational parsimony by serving most queries from one resident
checkpoint with a compact bank and auditable Trust Cards. Looking ahead, any
per-query router over merged checkpoints should pass a held-out statistical
test against the fixed merge before serving live traffic, and our certified
fallback provides that gate directly from out-of-fold predictions.

%% file: appendix.tex
\FloatBarrier
\section{Notation, Benchmarks, and Reproducibility}
\label{app:notation}

\Cref{tab:notation} collects the notation used throughout the paper.

\input{figures/v16_tab_notation.tex}

\subsection{Model Pairs, Benchmarks, and Truncation-Aware Scoring}
\label{app:datasets}

\paragraph{Checkpoint pairs.}
We evaluate four model pairs built from seven public checkpoints and one public
LoRA adapter. Phi-4 14B pairs Phi-4 with Phi-4-reasoning. Mutant 8B, our LoRA pair named
after its adapter, pairs Qwen3-8B with the same model after merging in the
rank-8 reasoning LoRA adapter prometechinc/Qwen3-8B-Mutant-Reasoning-Lora. Ministral 8B pairs
Ministral-3-8B-Base-2512 with Ministral-3-8B-Reasoning-2512. Qwen3-4B pairs
Qwen3-4B-Instruct-2507 with Qwen3-4B-Thinking-2507.

\paragraph{Benchmark scoring.}
Using a fixed random seed, we sample 100 truthfulness queries and 100 disjoint
calibration queries from TruthfulQA \citep{lin2022truthfulqa}.
Each truthfulness question is formatted as a four-option item combining the
reference best answer with three shuffled distractors, scored 1 when correct
and 0 otherwise. For calibration, the model also reports a confidence $c$ from
0 to 100, scored by $1-|c-\mathrm{correct}|$ with $c$ rescaled to $[0,1]$,
one minus a per-query form of the confidence-accuracy gap of
\citet{guo2017calibration}. Safety uses 100 XSTest prompts
\citep{rottger2024xstest}, split evenly between safe and unsafe requests,
scoring 1 for complying with a safe prompt or refusing an unsafe one. Every
action runs under either a strict JSON schema or a bare-answer format, with
unparseable outputs scoring 0, using greedy decoding up to 512 new tokens.

\paragraph{Content-based scoring at the generation limit.}
\label{app:scoring}
Every response reaching the generation limit is scored by its visible content
rather than naive refusal-phrase matching. We separate the visible response
appearing after the last closing think tag from the reasoning trace and treat
any unclosed trace as producing no answer. On safety prompts, a visible answer
counts as a refusal when it contains an explicit refusal statement or, on an
unsafe prompt, rebuts the harmful premise or refers the user to crisis help,
and counts as compliance when it provides a substantive answer of at least twenty
words. An unclosed trace never counts as compliance and scores 0 on
truthfulness and calibration. Across 254 human-annotated responses, this rule
agrees with manual labels on 0.921 of items.
Because it scores only the text delivered within the budget, it gives a lower
bound on longer decoding budgets. The reasoning variants of Phi-4 and Qwen3-4B
hit the 512-token cap on 98.7\% and 84.0\% of strict-JSON prompts,
explaining their low standalone scores in \Cref{tab:utility}.

\subsection{Reproducibility}
\label{sec:repro}
A single CPU evaluation script computes all reported statistics from cached
per-query outputs and router predictions under fixed-seed five-fold
cross-validation, using 1,000 within-objective cluster-bootstrap resamples per
cell and 2,000 resamples for pooled confidence intervals.
Code, merge bank definitions, cached outputs and predictions, 254 manual
annotations, and reproduction scripts will be released.

\FloatBarrier
\section{Merge Bank, Router Pool, and Baseline Adaptations}
\label{app:bank}

\paragraph{Candidate construction and fold-rebuilt bank validation.}
Candidate merges come from three sources: a $4\times4$ attention--FFN grid over
$\{0,\frac{1}{3},\frac{2}{3},1\}$ with remaining groups at $0.5$, local
half-step refinement adding up to 12 neighbors around the top two grid points
of each objective, and an 18-point Latin-hypercube design across all five
coefficients including both released endpoints. Running both output formats
produces 68 to 116 candidate actions per pair, from which the 95\% coverage
rule (\Cref{eq:bank_union}) retains 34 to 46 actions (\Cref{tab:bank}).

To verify that building the shared bank once from all benchmark queries
introduces no meaningful optimism, we also rebuild the bank strictly inside
each training fold. Under fold-rebuilt banks, the per-query oracle shifts
slightly from 88.92 to 88.42 points,
the rebuilt bank's fixed merge scores 0.06 to 0.22 points below $\afix(\w)$,
and PaSM stays level with that rebuilt fixed merge on all four fold
assignments, with every 95\% interval of its gain covering zero.

\input{figures/v16_app_bank.tex}

\paragraph{Router pool and must-route configurations.}
\label{app:ledger}
All bank routers read the base model's hidden state at layer 8, 16, or 24
averaged over final prompt tokens, with some designs adding TF-IDF features.
Across all designs and settings, we trained 195 configurations (41 to 60 per
pair). After within-pair Bonferroni correction, none of the 195 achieves a mean gain over $\afix(\w)$ significantly
above zero when forced to route, while 81 score significantly below it. \Cref{tab:ledger} lists the PaSM pool of 14
to 16 routers per pair selected alphabetically without inspecting results.

\input{figures/v16_app_pool.tex}

\subsection{Adaptation Fidelity and Native Baselines}
\label{app:proxydetails}

Every adapted baseline retains its method's core learned input-conditioned
router over the exact same 34 to 46 bank actions, folds, hidden-state features,
and statistics as PaSM (\Cref{tab:fidelity}), isolating routing mechanisms over
identical candidates. Components operating outside a fixed bank, such as
per-layer task-vector routing, test-time entropy training, BD-Merging's
closed-set evidential head, and Mediator's base-model fallback, are omitted in
the shared-bank adaptation; Mediator's fallback and WEMoE's per-layer routing are evaluated in their original form below.

\input{figures/v16_app_fidelity.tex}

\paragraph{Mediator with its native fallback.}
\label{app:ownfallback}
Retaining Mediator's uncertainty fallback to $\bbase$ whenever no action
reaches predicted probability 0.5 still leaves it 6.26 points below $\afix(\w)$
with a 95\% interval of $[-8.07,-4.44]$.
Redirecting that same confidence rule to $\afix(\w)$ without held-out
certification still trails by 3.94 points, because router confidence fails to
identify erroneous picks. Wrapping Mediator with our certified fallback holds
it at the fixed merge's utility with zero significantly worse cells.

\paragraph{Unadapted input-conditioned methods.}
\label{app:nativewemoe}
Evaluated in its original unadapted form on Qwen3-4B, where it routes over
per-layer FFN task vectors with its router's weights averaged over the prompt
tokens, WEMoE does
not significantly beat $\afix(\w)$, posting a mean gain of $+1.34$ points over
the 28 cells with a 95\% interval of $[-0.82,+3.52]$.
We also evaluate DaWin \citep{oh2025dawin}, which sets a per-input
interpolation weight from endpoint output entropies. On Qwen3-4B using the
authors' formula and mixture recipe, DaWin assigns nearly identical
coefficients across inputs of a given objective and format (standard deviation
at most 0.043), behaving like a single static merge with a non-significant gain
of $+0.33$ points and a 95\% interval of $[-1.16,+1.76]$.

\FloatBarrier
\section{Detailed Diagnostics on Must-Route Failure}
\label{app:mustroute}

\paragraph{Shared errors and participation ratio.}
\label{app:errors}
Across pair--objective blocks, 5 to 15\% of truthfulness queries and 1 to 8\%
of safety queries are unfixable by any bank action, exceeding a permuted-query
independence null in all 8 blocks. On queries where $\afix(\w)$ errs, each
other bank action succeeds 0.22 of the time on average versus 0.69 under
independence. Across all 12 pair--objective blocks, the participation ratio
$\mathrm{PR}=(\sum_i\mu_i)^2/\sum_i\mu_i^2$ of centered action
utilities lies between 1.98 and 4.74 for 34 to 46 actions, compared with 23.6
to 31.4 under independent errors.

Comparing equal-sized pools of eight models on identical strict-JSON queries
confirms that this collapse belongs to the merge path rather than shared item
difficulty. Across 500 random draws, eight actions from a single pair's bank
yield PR between 1.48 and 3.43, compared with 3.23 to 4.70 for the eight
distinct endpoint checkpoints across our four pairs, most visibly on
truthfulness and calibration.

\paragraph{Repair--break rates and choice-cost decomposition.}
In \Cref{eq:decomp}, the non-routable share $1-p$ lies between 0.81 and 0.91
across truthfulness and safety for must-route routers.
Breaking even on truthfulness would require adapted routers to achieve a repair
rate $r^\star=(1-p)\,e/p$ of 1.12 to 1.21, above the upper bound of one,
while they achieve 0.34 to 0.37 and direct 78.6 to 91.8\% of
truthfulness decisions to a single action on every pair. Even under
hypothetical independent errors at identical per-action accuracies, the
truthfulness routable share rises only from 0.110 to 0.198, still requiring a
break-even repair rate of 0.64 to 0.69 against the 0.34 to 0.37 achieved.

\Cref{tab:si} reports the full decomposition $\Delta_k=S+I$ for all six
must-route families. Choice cost $S$ is strictly negative in every entry,
reaching $-8.60$ to $-10.12$ points on truthfulness for adapted routers and
$-30.87$ for two-endpoint logistic regression, while per-query information $I$
reaches at most $+2.06$ points for any family.

\begin{table}[!htbp]
\centering
\footnotesize
\caption{\textbf{Leaving the fixed merge costs every must-route family more
than its per-query information recovers.} Gain over $\afix(\w)$ in points of
truthfulness or safety accuracy, weighted by $w_k$ and pooled over the pairs,
split into choice cost $S$ and information $I$.}
\label{tab:si}
\begin{tabular}{@{}lrrrr@{}}
\toprule
 & \multicolumn{2}{c}{Truthfulness} & \multicolumn{2}{c}{Safety} \\
\cmidrule(lr){2-3}\cmidrule(l){4-5}
Router family & $S$ & $I$ & $S$ & $I$ \\
\midrule
Mediator & $-10.12$ & $+1.78$ & $-1.68$ & $+1.20$ \\
WEMoE & $-8.60$ & $+0.02$ & $-0.66$ & $+0.17$ \\
BD-Merging & $-9.60$ & $+0.04$ & $-0.84$ & $-0.15$ \\
\hthree{} & $-9.78$ & $+0.90$ & $-0.60$ & $+0.05$ \\
Two-endpoint $k$NN & $-2.05$ & $+0.34$ & $-6.34$ & $-0.06$ \\
Two-endpoint log-reg & $-30.87$ & $+2.06$ & $-20.00$ & $-0.30$ \\
\bottomrule
\end{tabular}
\end{table}

\paragraph{Favorable training interventions.}
\label{app:rescue}
\Cref{tab:rescue} summarizes retraining under three favorable changes:
content-rule labels matching evaluation (leaving adapted routers 5.31 to 5.90
points below $\afix(\w)$), oracle objective identities evaluated against $a_k$
(5.83 to 6.33 points below),
and 1,113 Qwen3-4B queries representing 3.3 to 4.5 times the original training
set. With extra data, adapted routers beat $\afix(\w)$ by 4.20 to 9.71 points
on training folds yet still lose by 1.94 to 2.54 points out of sample.

\begin{table}[!htbp]
\centering
\footnotesize
\setlength{\tabcolsep}{6pt}
\caption{\textbf{Evaluation-matched labels, oracle objective identities, and
more data do not rescue a must-route router.} Gain in points. Matched labels:
retrained on content-rule labels against $\afix(\w)$ over 112 cells.
Told objective: top action for each query's true objective against $a_k$ over
112 cells.
More data: adapted routers retrained on 3.3 to 4.5 times more Qwen3-4B queries
against $\afix(\w)$ over its 28 cells.}
\label{tab:rescue}
\begin{tabular}{@{}lrrr@{}}
\toprule
Router family & Matched labels & Told objective & More data \\
\midrule
Mediator & $-5.31$ & $-6.33$ & $-2.54$ \\
WEMoE & $-5.88$ & $-5.99$ & $-1.94$ \\
BD-Merging & $-5.90$ & $-5.83$ & $-2.11$ \\
\hthree{} & $-5.63$ & $-6.12$ & $-2.07$ \\
Two-endpoint $k$NN & $-2.11$ & $-6.78$ & --- \\
Two-endpoint log-reg & $-10.87$ & $-14.51$ & --- \\
\bottomrule
\end{tabular}
\end{table}

\FloatBarrier
\section{Statistical Level, Power, and Robustness Checks}
\label{app:theory}

\paragraph{Significance level and Bonferroni control.}
For a single router $\pi$ fixed prior to validation, admitting $\pi$ when
$L_{\w}(\pi)>0$ (\Cref{eq:boot_bound}) implements a one-sided
percentile-bootstrap test of $G_{\w}(\pi)\le 0$ at nominal level $\alpha=5\%$,
which is asymptotically exact as validation size grows. Because standard
out-of-fold predictions come from routers whose training folds include the
outer test fold, the nominal level is approximate; nested selection below
removes this dependence for every router except the re-merge router. Applying a
Bonferroni correction across the pool of 14 to 16 routers at level
$\alpha/|\Pi|$ with 10,000 resamples yields a PaSM gain of $+0.06$ points with
a 95\% interval of $[-0.03,+0.14]$ and zero significantly worse cells.

\paragraph{Synthetic and planted-gain power.}
Pooled across a synthetic gain grid, a router with true gain at or below $-4$
points is admitted in 0.005\% of decisions, one between $-1$ and $0$ points in
0.59\%, and one gaining 8 to 12 points in 97\% of decisions while retaining
95\% of its gain.
\Cref{tab:e17power} likewise tests a planted-gain router that outputs the best
bank action on a random share $\rho$ of queries and WEMoE's choice elsewhere.
Admission climbs to 55.9\% at $\rho=0.5$ and 99.8\% at $\rho=1$, while in the
losing regime ($\rho\le 0.2$) the fallback yields small positive gains of
$+0.04$ to $+0.29$ points.

\begin{table}[!htbp]
\centering
\footnotesize
\caption{\textbf{The fallback admits a router more often as its gain grows and
incurs no net loss while the router loses.} Planted-gain router: bank oracle on
a share $\rho$ of queries, WEMoE elsewhere; means over 20 replicates.}
\label{tab:e17power}
\begin{tabular}{@{}lrrr@{}}
\toprule
$\rho$ & Planted gain & Admitted (\%) & Gain with fallback \\
\midrule
0 & $-5.91$ & 1.6 & $+0.04$ \\
0.05 & $-4.89$ & 2.3 & $+0.08$ \\
0.1 & $-3.89$ & 3.6 & $+0.14$ \\
0.2 & $-1.54$ & 8.3 & $+0.29$ \\
0.3 & $+0.70$ & 18.8 & $+0.72$ \\
0.5 & $+5.07$ & 55.9 & $+3.58$ \\
0.75 & $+10.39$ & 90.8 & $+9.81$ \\
1 & $+15.54$ & 99.8 & $+15.51$ \\
\bottomrule
\end{tabular}
\end{table}

\subsection{Robustness Across Splits, Budgets, and Unseen Queries}
\label{app:robustness}

\paragraph{Multiple fold assignments.}
\label{app:foldseeds}
\Cref{tab:utility} averages four independent random fold assignments, each
refitting fixed merges, certification, and all routers except 29 configurations
(including the two re-merge routers) that keep their original held-out
predictions.
On every assignment, standalone adapted routers degrade 50 to 72 of the 112
cells, while wrapped families degrade at most 2 (\Cref{tab:foldseeds}).
Across the four assignments, PaSM leads $\afix(\w)$ on safety-dominant
preferences by $+0.35$, $+0.47$, $+1.07$, and $+0.40$ points (mean $+0.57$),
and certification improves each adapted router's worst cell from $-19.88$ to
$-23.62$ points alone to $-1.00$ to $-2.56$ behind the fallback.

\begin{table}[!htbp]
\centering
\small
\caption{\textbf{On every fold assignment, standalone adapted routers degrade
at least 50 of 112 cells, whereas wrapped families degrade at most 2.} Mean
utility of $\afix(\w)$ and PaSM gain over $\afix(\w)$ in points; counts of
significantly worse cells out of 112.}
\label{tab:foldseeds}
\begin{tabular}{lcccc}
\toprule
Fold assignment & 1 & 2 & 3 & 4 \\
\midrule
$\afix(\w)$, mean utility & 73.38 & 73.77 & 73.32 & 72.94 \\
PaSM, gain & $-0.02$ & $+0.08$ & $+0.52$ & $+0.14$ \\
Four adapted routers, worse cells & 55--61 & 62--72 & 56--69 & 50--60 \\
Six families behind the fallback, most worse cells & 2 & 0 & 0 & 0 \\
\bottomrule
\end{tabular}
\end{table}

\paragraph{Nested cross-validation selection.}
\label{app:nested}
Setting each outer test fold aside, retraining every router except the re-merge
router by inner five-fold cross-validation on the remaining four folds,
selecting $\afix(\w)$ on inner training folds, and running \Cref{alg:fallback}
strictly on inner predictions removes cross-fold training overlap. Every
wrapped router family and PaSM (mean gain $+0.03$ points) maintain parity with
$\afix(\w)$.

\paragraph{Longer generation budget.}
\label{app:budget}
Re-evaluating all 39 Qwen3-4B bank actions and both endpoints on 300 queries at
4,096 new tokens drops truncation from 21.8 to 30.8\% per objective down to 0.9
to 2.0\%. Under this longer budget, adapted routers lose even more, trailing by
4.77 to 5.98 points versus 2.52 to 3.65 at 512 tokens, while every wrapped
router stays within 0.07 points of $\afix(\w)$ with zero significantly worse
cells, and PaSM gains $+0.04$ points.

\paragraph{Unseen held-out queries.}
\label{app:confirm}
Scoring all bank actions on 813 newly drawn Qwen3-4B queries (232 truthfulness,
350 XSTest safety, and 231 calibration) without refitting the bank or routers,
with two routers rebuilt from the original ones, shows that PaSM maintains
parity at $-0.06$ points with a 95\% interval of $[-0.16,+0.04]$ and zero
significantly worse cells.

%% file: figures/v16_tab_notation.tex
\begin{table}[!htbp]
\centering
\small
\caption{\textbf{Notation.}}
\label{tab:notation}
\setlength{\tabcolsep}{4pt}
\begin{tabular}[t]{@{}>{\raggedright\arraybackslash}p{\dimexpr\textwidth/8\relax}>{\raggedright\arraybackslash}p{\dimexpr\textwidth*3/9\relax}@{}}
\toprule
\textbf{Name} & \textbf{Meaning} \\
\midrule
$\bbase,\,\bint$ & base checkpoint and specialized variant of a pair \\
$\blam\in[0,1]^5$ & merge vector across five parameter groups (\S\ref{sec:problem}) \\
$\bank$ & bank: finite set of actions (merge and answer format) on one merge path \\
$\w$ & preference weights $(w_{\mathrm{tr}},w_{\mathrm{sf}},w_{\mathrm{cal}})$ on the simplex \\
$\pi$;\ $\Pi$;\ $f$ & router; candidate router pool; held-out test fold \\
$\Util_{k}(q,a)$ & utility in $[0,1]$ of action $a$ on query $q$ of objective $k\in\{\mathrm{tr},\mathrm{sf},\mathrm{cal}\}$ \\
$V_{\w}(\pi)$ & preference-weighted utility of router $\pi$ (\Cref{eq:objective}) \\
$\afix(\w)$ & per-preference fixed merge: best bank action for $\w$ on training queries \\
$a_k$ & per-objective fixed merge: best bank action for objective $k$, chosen out of fold \\
$h_{k}(q)$ & routing headroom of query $q$ (\Cref{eq:headroom}) \\
$G_{\w}(\pi)$;\ $\hat{G}_{\w}(\pi)$ & true and out-of-fold gain over $\afix(\w)$ (\Cref{eq:oof_gain}) \\
$L_{\w}(\pi)$;\ $\alpha$ & bootstrap lower bound at level $\alpha$ (\Cref{eq:boot_bound}) \\
\bottomrule
\end{tabular}\hfill
\begin{tabular}[t]{@{}>{\raggedright\arraybackslash}p{\dimexpr\textwidth/8\relax}>{\raggedright\arraybackslash}p{\dimexpr\textwidth*3/9\relax}@{}}
\toprule
\textbf{Name} & \textbf{Meaning} \\
\midrule
$\mathcal{C}_{\w}$;\ $\pi^{\star}$ & certified router set ($L_{\w}(\pi)>0$) and top certified router (\Cref{eq:pasm_policy}) \\
$\mathcal{T}(q,\w)$ & per-request Trust Card recording the served path and certificate (\S\ref{sec:trust_card}) \\
cell & one pair under one preference $\w$ \\
must-route & a router with no fallback to $\afix(\w)$ \\
parity & mean gain of at least $-0.10$ points with at most 4 significantly worse cells (\S\ref{sec:setup}) \\
$p,\,r$ & routable share; share of it a router repairs \\
$1-p,\,e$ & rest share, where no repair is possible; share of it a router breaks \\
$\Delta_{k}$ & gain on objective $k$ as a share of its queries (\Cref{eq:decomp}) \\
$g_{\mathrm{out}}$ & gain from actions outside the bank \\
$r^\star$ & repair rate needed to break even, $(1-p)\,e/p$ (\S\ref{sec:decomposition}) \\
$S,\,I$ & choice cost (signed utility penalty when negative) and information term, $\Delta_{k}=S+I$ (\Cref{eq:choice_cost}) \\
\bottomrule
\end{tabular}
\end{table}

%% file: figures/v16_app_bank.tex
\begin{table}[!htbp]
\centering
\footnotesize
\caption{\textbf{The bank keeps 34 to 46 actions per pair, a third to a half of
the candidates.} An action is a merge in one of two answer formats. Kept per objective:
actions that cover 95\% of the objective's per-query gain over the default
action; the bank is their union, so objectives share actions. Headroom $\geq5$: percentage of
queries on which some bank action beats the per-objective fixed merge by at
least 5 points (\Cref{eq:headroom}).}
\label{tab:bank}
\setlength{\tabcolsep}{2.5pt}
\begin{tabular}{@{}llrrrrccccccc@{}}
\toprule
 & & \multicolumn{2}{c}{\textbf{Candidates}} & \multicolumn{2}{c}{\textbf{Bank}} &
\multicolumn{3}{c}{\textbf{Kept per objective}} &
\multicolumn{3}{c}{\textbf{Headroom $\geq5$}} \\
\cmidrule(lr){3-4}\cmidrule(lr){5-6}\cmidrule(lr){7-9}\cmidrule(l){10-12}
\textbf{Pair} & \textbf{Variant} & Merges & Actions & Actions & Merges & Tru. & Saf. & Cal. & Tru. & Saf. & Cal. \\
\midrule
Phi-4 14B    & Phi-4-reasoning & 34 & 68  & 34 & 27 & 12 & 9  & 28 & 16 & 13 & 64 \\
Mutant 8B    & Qwen3-8B + Mutant LoRA & 58 & 116 & 40 & 40 & 28 & 23 & 34 & 14 & 15 & 36 \\
Ministral 8B & Reasoning-2512 & 46 & 92  & 46 & 32 & 19 & 18 & 38 & 12 & 21 & 63 \\
Qwen3-4B     & Thinking-2507 & 46 & 92  & 39 & 32 & 8  & 12 & 32 & 6  & 23 & 52 \\
\bottomrule
\end{tabular}
\end{table}

%% file: figures/v16_app_pool.tex
\begin{table}[!htbp]
\centering
\footnotesize
\caption{\textbf{The PaSM pool keeps one router per design, 14 to 16 per pair,
and every router enters the certified fallback the same way.} The uncertainty-abstention, two-stage, and
committee routers lean toward $\afix(\w)$ on uncertain queries. Pairs: Mu Mutant, Mi Ministral, Q Qwen3-4B, P Phi-4.}
\label{tab:ledger}
\setlength{\tabcolsep}{4pt}
\renewcommand{\arraystretch}{1.08}
\begin{tabular}{@{}>{\raggedright\arraybackslash}p{0.2\textwidth}>{\raggedright\arraybackslash}p{0.56\textwidth}p{0.1\textwidth}@{}}
\toprule
\textbf{Design} & \textbf{What it does} & \textbf{Pairs} \\
\midrule
Nearest neighbor & $k$NN over TF-IDF and projected hidden features, predicting per-action utility & all \\
Best-action classifier & per-objective classifier of the best action & all \\
Feed-forward net & per-objective regression of every action's utility & all \\
Distilled shared model & shared trunk with action-geometry heads, trained on soft utility targets & all \\
Deep ensemble & five independently seeded distilled models, averaged & all \\
Beta hypernetwork & low-rank, query-dependent perturbation of an action-scoring matrix & all \\
Energy model & energy over merge coefficients & all \\
Token attention & attention pooling over token states & Mu, Mi \\
Uncertainty abstention & MC-dropout; abstains on the 30\% most uncertain queries & all \\
Two-stage & routes a query when its predicted headroom $\hat h(q)$ exceeds 0.05 & Mi, Q, P \\
Committee & nearest neighbor, distilled model, and feed-forward net; abstains when they disagree & all \\
WEMoE, Mediator, BD-Merging, \hthree{} & routing mechanisms of these methods over bank actions & all \\
Re-merge & a feed-forward net predicts attention and FFN coefficients; the merge is built for the query & Mu, Mi \\
\bottomrule
\end{tabular}
\end{table}

%% file: figures/v16_app_fidelity.tex
\begin{table}[!htbp]
\centering
\footnotesize
\caption{\textbf{Every adapted router keeps its method's input-conditioned router and
runs it over the same bank actions, folds, features, and statistics as PaSM.}
Each cell gives the status of the original component in our adaptation and,
after the arrow, what replaces it. Preference handling is \textsc{added}: none
of the four originals takes a preference. The dropped components are the
per-layer and uncertainty machinery that operates outside a fixed bank and \hthree{}'s three aligned experts.}
\label{tab:fidelity}
\setlength{\tabcolsep}{3pt}
\renewcommand{\arraystretch}{1.08}
\newcommand{\fK}{\textsc{kept}}
\newcommand{\fC}{\textsc{changed}}
\newcommand{\fD}{\textsc{dropped}}
\newcommand{\fA}{\textsc{added}}
\begin{tabular}{@{}>{\raggedright\arraybackslash}p{0.13\textwidth}*{4}{>{\raggedright\arraybackslash}p{0.198\textwidth}}@{}}
\toprule
\textbf{Component} & \textbf{WEMoE} & \textbf{Mediator} & \textbf{BD-Merging} & \textbf{\hthree{}} \\
\midrule
Input-conditioned router & \fK: router over hidden features & \fK: learned score over candidates & \fK: per-sample router on base features & \fK: gated mixture of experts \\
Granularity & \fK: per input & \fK: per sample & \fK: per sample & \fC: per token and layer $\to$ per input \\
Router input & \fC: each layer's FFN input $\to$ base hidden state, layer 16 & \fC: base early layers $\to$ base hidden state, layer 16 & \fC: base last layer $\to$ base hidden state, layer 16 & \fC: each layer's token state $\to$ base hidden state, layer 16 \\
Router output & \fC: task-vector weights $\to$ one bank action & \fC: top-$k$ task experts $\to$ one bank action & \fC: task-vector weights $\to$ one bank action & \fC: FFN expert outputs $\to$ mixed action scores \\
Training signal & \fC: test-time entropy $\to$ best action per query & \fC: task classifier $\to$ best action per query and preference; task head kept as auxiliary & \fC: entropy and contrastive loss $\to$ best action per query & \fC: language-model and gating loss $\to$ best action per query \\
Per-layer merging & \fD: per-layer FFN task-vector routing & \fD: layer-wise conflict split & \fD: layer-wise weights & \fD: per-layer FFN experts \\
Uncertainty or conflict & none in the original & \fD: temperature, top-$k$, base-model fallback & \fD: evidential head, neighbour discrepancy & \fC: expert drift penalty $\to$ coefficient drift penalty \\
Experts & \fC: task vectors $\to$ bank actions & \fC: task experts $\to$ bank actions & \fC: task vectors $\to$ bank actions & \fD: three aligned models; heads not tied to an objective \\
Preference & \fA: weights the objective routers & \fA: router input & \fA: weights the objective routers & \fA: weights the objective routers \\
\bottomrule
\end{tabular}
\end{table}